\documentclass[twocolumn,10pt]{asme2e}

\usepackage{graphicx}
\graphicspath{ {images/} }
\usepackage{algorithmic}
\usepackage{cite}
\usepackage{subfigure}
\usepackage{array}
\usepackage{afterpage}
\usepackage{amsfonts}
\usepackage{mathtools}
\usepackage{amsmath}
\usepackage{amssymb}
\usepackage{relsize}
\usepackage{mathrsfs} 
\usepackage[dvipsnames]{xcolor}
\usepackage{todonotes}
\usepackage{color}
\usepackage{ulem} 						
\usepackage{xspace}

\usepackage{parskip}
\usepackage{color}
\usepackage[unicode=true,
    pdfborder={0 0 0},
    colorlinks=false,
    bookmarks=true,
    bookmarksopen=true,
    bookmarksopenlevel=2,
    bookmarksnumbered=true,
    pdfpagelayout=SinglePage,
    hyperfootnotes=false,
    pdfpagemode=UseNone,
    pdftoolbar=false,
]{hyperref}

\confshortname{IDETC/CIE 2018}
\conffullname{The ASME 2018 International Design Engineering Technical Conferences \&\\
              Computers and Information in Engineering Conference}

\confdate{26-29}
\confmonth{August}
\confyear{2018}
\confcity{Quebec}
\confcountry{Canada}

\papernum{DETC2018-86359}

\title{Draft: An Integrated Design and Simulation Environment for Rapid Prototyping of Laminate Robotic Mechanisms}
\author{Mohammad Sharifzadeh\thanks{Authors contributed equally.}
    \affiliation{
	IDEALAB\\
	The Polytechnic School\\
	Ira A. Fulton Schools of Engineering\\
    Arizona State University\\
	Mesa, Arizona 85212\\
    Email: msharifz@asu.edu
    }	
}

\author{Roozbeh Khodambashi$^{*}$\thanks{Address all correspondence to this author.}
    \affiliation{IDEALAB\\
	The Polytechnic School\\
	Ira A. Fulton Schools of Engineering\\
    Arizona State University\\
	Mesa, Arizona 85212\\
    Email: rkhodamb@asu.edu
    }
}
\author{Daniel M. Aukes 
    \affiliation{IDEALAB\\
	The Polytechnic School\\
	Ira A. Fulton Schools of Engineering\\
    Arizona State University\\
	Mesa, Arizona 85212\\
    Email: danaukes@asu.edu
    }
}

\begin{document}

\maketitle    

\begin{abstract}
{\it Laminate mechanisms are a reliable concept in producing low-cost robots for educational and commercial purposes. These mechanisms are produced using low-cost manufacturing techniques which have improved significantly during recent years and are more accessible to novices and hobbyists. However, iterating through the design space to come up with the best design for a robot is still a time consuming and rather expensive task and therefore, there is still a need for model-based analysis before manufacturing. Until now, there has been no integrated design and analysis software for laminate robots. This paper addresses some of the issues surrounding laminate analysis by introducing a companion to an existing laminate design tool that automates the generation of dynamic equations and produces simulation results via rendered plots and videos. We have validated the accuracy of the software by comparing the position, velocity and acceleration of the simulated mechanisms with the measurements taken from physical laminate prototypes using a motion capture system.}
\end{abstract}



\section{INTRODUCTION}
\begin{figure*}[t]
\centering
\includegraphics[scale=0.28]{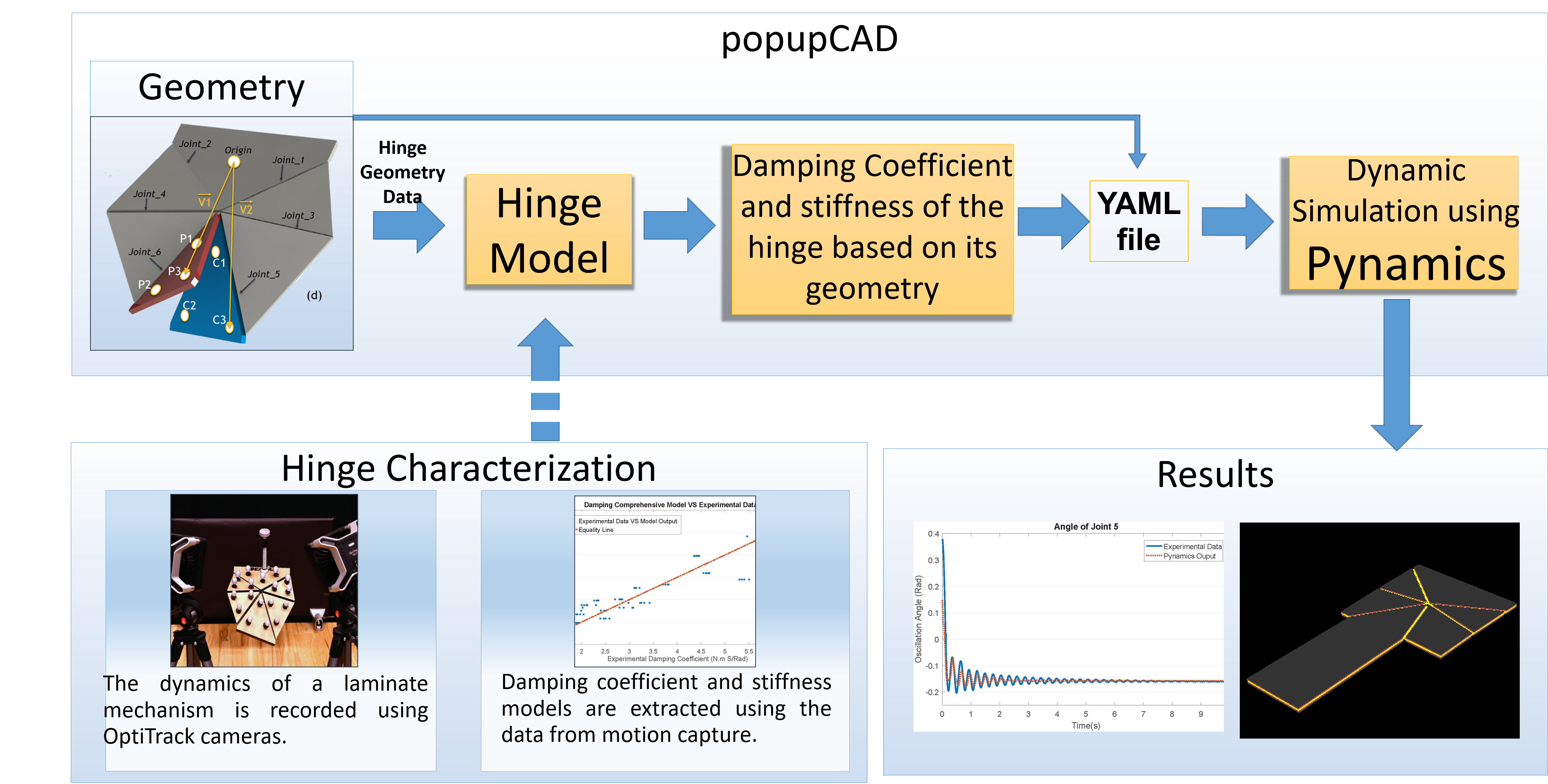}
\caption{The internal work flow of for design, dynamics simulation and manufacturing of a laminate mechanism. popupCAD is used for creating the geometry and manufacturing files needed for a laser cutter. Pynamics is a dynamic analysis tool integrated into popupCAD. The geometry data along with the damping coefficient and stiffness obtained from experiments on the hinges are combined in a YAML file which is read by Pynamics for analyzing the dynamics behavior. }
\label{pic_workflow}
\end{figure*}

Robotics is a difficult field for non-experts to enter, as traditional robots are expensive to purchase and require a large amount of analytical skill and technical expertise to design, build, and field successfully.  Recently, advances in lower-cost fabrication techniques have made it easier for novices to prototype parts and mechanisms quickly and easily using laminate techniques. These techniques have also made it possible for new generations of millimeter-scale, lightweight, and low-cost robotic mechanisms to be prototyped with ease\cite{Baisch2014,Ma2013a}. The mechanisms produced with these methods typically feature flexure hinges -- composed of Polyimide, Polyester, fabric, etc. -- embedded in rigid laminate bodies, which, when exposed at joints, make it possible to create precise mechanisms that rely on material deformation to define the stiffness, damping, and thus, motion of the system. 

While manufacturing knowledge has advanced due to new Computer-Aided Manufacturing(CAM) tools\cite{Aukes2014,Aukes2014a}, analyzing the motion and performance of such devices has lagged. This often involves understanding the dynamics involved with operating a robot in an unstructured environment -- an advanced technique unavailable to many.

The precision associated with fabrication and the assumption of pin-joint hinges permits many of these structures to be approximated as traditional mechanisms; a variety of work has been done for specific mechanisms to understand resulting kinematics of muti-bar closed-loop systems\cite{Aukes2014c}. However, material selection plays a significant role in deflection of these devices, as material bending at hinges can influence the stiffness and damping of the system as a whole\cite{Stellman2005}. Some approaches have used structural engineering methods to understand system stiffness and to solve the static force balances that these structures can accommodate given external loading\cite{Tachi2009,Schenk2011,Fuchi2014}, while others have looked at higher-order models for flexure hinges\cite{Ma2013}. These approaches are useful for understanding the quasi-static deformation, as well as the linearized system stiffness of a given configuration.

Due to inertia, high velocities, and intrinsic damping and stiffness present in the joints, many of the devices being made must be considered dynamic.  Kinematic and stiffness-based solutions are not sufficient to understand device motion.  While deriving the equations of motion for laminates has been performed for specific devices \cite{Doshi2015c,Hanna2014}, less has been done to use the properties of laminate systems in order to understand and solve for the motion of devices in general. This is due to several reasons.  First, since they are manufactured in a flat state, laminate devices begin in an  inherently singular configuration and must be erected into a valid 3D shape on one side or another of a singularity, as discussed in \cite{Hanna2014}. Due to these singularities, laminate mechanisms often have multiple potentially-valid configurations, which must be specified by the user or guessed by an automated system.  In addition, laminate mechanisms often form parallel chains of multiple links.  Such loop constraints are difficult to specify in general, while maintaining valid and consistent initial conditions across singularities.   
 
In previous work we introduced popupCAD\cite{Aukes2014,Aukes2014a}, a design environment which automates the computations required to manufacture laminate devices\footnote{\url{http://www.popupcad.org/}}.  This tool permits object-oriented design methodologies and considers the constraints of laminate fabrication processing steps in order to produce manufacturable laminate cut files.  We have also introduced the addition of a dynamic and a finite element analysis tool for use within the popupCAD development environment \cite{Aukes2015}, however we did not address the main challenges associated with simulation as mentioned above. Moreover, we did not integrate these tools in the main popupCAD software and therefore, the user has to customize the simulation process for each design. This requires the user to spend more time on training which slows down the prototyping process.

The contribution of this paper is two-fold. First, it presents an extension to popupCAD which permits the generation of the equations of motion directly from popupCAD geometries. This integrated design environment, which has been described in Fig.~\ref{pic_workflow} offers a general method for design and analysis of laminate devices with minimum user interference required which makes it possible to automatically determine the motion of laminate mechanisms with closed-loop kinematics while simultaneously solving for valid initial conditions.  Second, it introduces a general guideline for characterizing the hinges used in laminate devices which can then be used by other researchers to study other mechanisms made of hinges with different materials and geometries. 

This paper is organized as follows: 
in Section~\ref{sec:model-generation}, we describe the fundamentals of the popupCAD simulation environment along with some of specific modeling decisions made and challenges solved. 
Since the damping coefficient and stiffness of the hinge material are critical to a successful dynamic simulation, Section~\ref{sec:hinge-characterization} introduces an experimental procedure for obtaining dynamic parameters for the hinges in a simple system and introduces methods of extrapolating the results to new hinge designs. In Section~\ref{sec:results}, we discuss the results of hinge characterization experiments and also present simulation and experimental results on arbitrary laminate mechanism designs.
Finally in Section~\ref{sec:conclusions} we conclude with thoughts on next steps and future work.



\section{DYNAMIC SIMULATION IN POPUPCAD}\label{sec:model-generation}

\begin{figure}[t]
\centering
\includegraphics[scale=0.35]{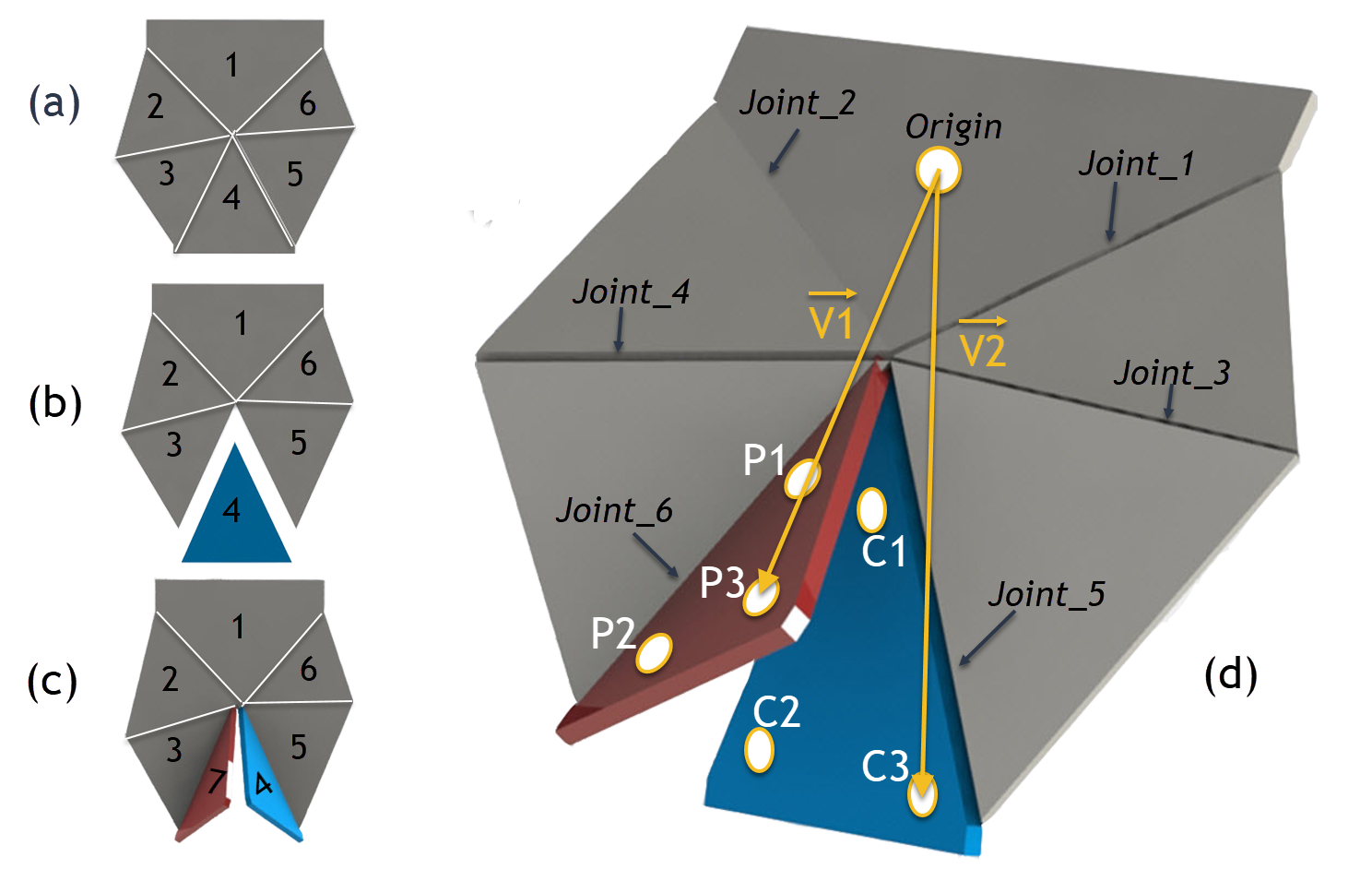}
\caption{The schematic showing the steps for mathematically defining the closed-loop constraints.}
\label{pic_baumgarte}
\end{figure}
We have addressed the issues mentioned in the previous section by introducing a new suite of Python-based tools for simulating rigid-body dynamics in laminates\footnote{\scriptsize\url{https://github.com/idealabasu/code_foldable_dynamics}}.   This functionality is designed to work with several other tools we have previously developed, specifically popupCAD and Pynamics\footnote{\scriptsize\url{https://github.com/idealabasu/code_pynamics.git}}, a symbolic toolkit for generating equations of motion.  Pynamics is particularly useful for several reasons.  It can describe vectors using symbolic variables, take time derivatives of vectors in multiple reference frames, and can use Kane's method to derive equations of motion symbolically.  This gives more insight about mechanism motion than numerical methods because one can see the contribution of individual parameters to the evolution of each state variable. 

To merge the capabilities of popupCAD and Pynamics, we have made the following additions.  First, we have made it possible to read popupCAD designs and extract rigid body information, as well as detailed joint information, making it possible to simulate a device drawn in popupCAD. Second, we have added the capability to detect and handle open and closed-loop mechanisms automatically, a necessity for laminate mechanisms. We have addressed strategies for determining valid initial conditions.  And finally, we have made it possible to render the motion of laminate bodies directly to video or animate 3D motion within a Python-based GUI.  Together, these innovations make it possible for novice designers to simulate a wide variety of laminate mechanisms and visualize the motion of their device during the design phase, rather than after prototyping.

\subsection{Importing from popupCAD}
Pynamics receives geometry data and material properties of laminate layers from popupCAD as a YAML (YAML Ain't Markup Language) file. YAML is a human friendly data serialization
  standard for all programming languages.  Rigid bodies, hierarchical interconnections of bodies via rotational joints, fixed (Newtonian) bodies and  material properties are stored in this file as Python-based classes after the YAML file is read by Pynamics. A hierarchical tree represents the network of connected mechanisms in which rigid bodies are the nodes and the joints connecting them are branches of the tree. Trees are good at representing serial chains of bodies, but do not adequately capture the topology of parallel mechanisms.

For such devices, we discuss adding closed-loop constraints in Section \ref{ssec:c}.
\subsection{Obtaining the Dynamic Model}
There are a variety of ways to describe the kinematics and dynamics of a physical system. To minimize the size of our system, we have selected a reduced-coordinate system with state variables corresponding to joint rotation, which is described in this section.

A reference frame is created for each body to represent its orientation in space with respect to its parent. The orientation of each body is defined with all axes initially aligned with the base frame. This frame is then rotated along a vector defined by two joint coordinates between the bodies, with $q$ defining the angular displacement between the two. To clarify this, consider Fig.~\ref{pic_baumgarte}(a) which shows a 6-bar laminate mechanism with each body numbered from 1 to 6. We use this mechanism as a motivating example throughout the paper due to its relative complexity, non-symmetric angles, and closed-loop topology. Body 1 is defined in popupCAD as the fixed Newtonian reference frame. In this representation, body 1 is considered as the parent and body 2 and 3 which are connected to it are considered as the children of this parent. Body 1 is at the top of the hierarchy which is also called the first generation. All bodies connected to body 1 are considered as the second generation. This hierarchy continues with one branch on the left which contains body 2 and 3, and another branch on the right which contains body 5 and 6. 
Body 4 is the last body in the hierarchy, which can be either connected to body 3 or 5 as shown in Fig.~\ref{pic_baumgarte} (b). We describe how  this body is treated in the next section while considering kinematic constraints.  	

\subsection{Adding kinematic constraints and initial conditions}\label{ssec:c}	
In creating a general and valid set of constraint equations for closed-loop laminate mechanisms, we should note that each joint may have associated dynamical elements such as springs and dampers defined by the flexure material, making it difficult to create a constraint along the joint. Our constraint generation instead relies on the fact that the same rigid body, taken from two different branches of the same kinematic loop, should occupy the same space. To represent this mathematically, we use dummy bodies. 
As a general example, consider the 5 bar mechanism with a single closed loop chain shown in Fig.~\ref{pic_baumgarte} c.
We create a dummy body 7 and attach it to body 3, assigning half of the mass and inertia of body 4 to it. The mass and inertia of body 4 should also be divided by half. In this way we have two rigid bodies attached to the ends of each serial chain that are meant to represent the same single body. We express the position of three points on body 4 and seven as
\begin{align}
\left(\vec{V_2}-\vec{V_1}\right) \cdot \left(\vec{V_2}-\vec{V_1}\right) = 0 \label{Eq_Baumgartes_Constraints},
\end{align}

where $\vec{V_2}$ and $\vec{V_1}$ represent the distance from the origin to two similar points $C_3$ and $P_3$ on body 4 and 7, respectively. This comes from the fact that the coordinates of any point on body 4 measured from the fixed reference frame should match the coordinates of the corresponding point on body 7 measured from the same reference frame. A total of three equations representing three non-co-linear points are needed to fully constrain the position and orientation of the bodies together. For multiple loops, the process should be repeated for each loop of the mechanism. However, in this paper we have considered mechanisms with only one closed loop, and cannot assume that our method will work without modification when extended to multi-loop kinematic chains. 
Next, we add the forces resulting from springs, dampers, gravity and external torques acting on the joints, using parameters supplied by the YAML file. 
Initial conditions are provided as a vector containing the relative angle between each pair of connected bodies ($q$) and their relative angular velocity ($\dot{q}$). Therefore, in the case of the 6-bar mechanism, the vector of initial conditions contains 12 elements. In Sec.~\ref{Sec_dd} we  describe our method for finding a valid set of initial conditions. 
\subsection{Integration and solving}
\label{Sec_dd} 

Since joints in laminate mechanisms are created via flexible material layers, laminate mechanisms typically emerge from fabrication in a singular configuration. After fabrication, an assembly step typically erects a flat laminate into a three-dimensional shape.  Resulting mechanisms are then typically constrained or operated in conditions which prevent singularities from occurring. However, for simulation purposes,  a set of valid, non-singular initial conditions must be determined in order to integrate. Using a traditional Lagrange formulation to constrain closed-loop mechanisms in the presence of initial-value singularities is a problem, because the introduction of any non-zero initial value in state variables produces a permanent error in position constraints. To reduce this error and to deal with singularities simultaneously, we use Baumgarte's method \cite{Baumgarte1972} further developed by Masarati \cite{masarati2011adding}.
Using constraint equations with Baumgarte's method eliminates invalid initial value guesses over successive simulation steps, with the $\alpha$ and $\beta$ terms (introduced in \cite{Baumgarte1972}) behaving like a second order system to minimize error over successive integrations. Constraint stabilizing is controlled by $\alpha$ and $\beta$-influenced terms, which simultaneously eliminate error and constrain the closed-loop mechanism.  $\alpha$ and $\beta$ were determined by trial and error for the 6-bar mechanism studied in this paper. For these parameters, approximately 300 time steps are needed for the constraint error to approach zero within desired tolerances. 

Simulation of the mechanism occurs in two steps. Again, using the example from Fig.~\ref{pic_baumgarte}  as a motivating example, an initialization step solves errors in constraints from an initial position guess supplied by the user.  
This may be a very rough guess about the value of one of the joint angles, as Baumgarte's method will produce a valid configuration after sufficient cycles. From the first step, 
the mechanism is then simulated for a number of time steps which is found experimentally to make sure all the errors become sufficiently small. During this time, the mechanism reaches a valid non-singular configuration under the influence of gravity, joint forces, and initial-value constraints. 

The final values of the joint angles from the first step are then fed into the second simulation and used as valid initial conditions to determine the motion of the mechanism. It should be noted that the mechanism is in equilibrium only when the external forces or torques are present. Removing these external forces will cause the mechanism to start moving. This is simulated in this second step. In this second simulation, Baumgarte's method is not used to increase simulation speed.  This is possible because initial error has been eliminated, and what little error gets subsequently introduced is negligible with respect to the valid configuration. This has the positive impact of increasing integration speed, while removing the impact of $\alpha$ and $\beta$ terms on the physical simulation.

\section{HINGE CHARACTERIZATION}\label{sec:hinge-characterization}
As mentioned earlier, in order to simulate a laminate mechanism, we need to insert the values for the damping coefficient and stiffness of the hinge material into popupCAD. These values are dependent on a variety of parameters such as material properties and the geometry of the hinges. Although these parameters can be arbitrarily inserted into the simulation, in order to get results that are comparable to the real-world situation, we need to have an estimate of the values of these properties which are close to the hinges used in actual laminate designs. Having a model that takes the material properties and geometry of a hinge as input and provides the damping coefficient and stiffness of the hinge as output would be of great help to the user since it is almost impractical to perform detailed tests to obtain these parameters for each design. This section describes studying a hinge in a simple pendulum made using laminate manufacturing techniques in order to map design variables to stiffness and damping in a parametric hinge joint which later can be used for extending simulations to new and more complex systems. Doshi et al \cite{Doshi2015c} have done similar work for retrieving joint parameters by using spring and damping coefficients using a standard second-order system, and applying these to the hand-coded dynamics of a given five-bar mechanism in a vacuum. The methods used in this paper work in air and extract model parameters using the dynamic model of our system described in Section \ref{sec:motion-capture}. In addition, this paper uses different material models in order to more closely match experimental data. A set of material tensile tests have also been performed to determine the allowable motion range of the hinges.
\subsection{Material test}
The 0.127\,mm-thick flexible polyester used to create flexure hinges was tested according to ASTM D882-12 standard for tensile properties of thin plastic sheeting\cite{ASTMInternational2012}. Specimens were prepared according to this standard and tightened between two smooth, hard-rubber jaws of the tensile testing machine since serrated grippers created resulted in premature failure of the specimen.
The test was repeated for 11 specimens and the average Young's modulus and yield stress obtained was 4383.27 MPa and 42.84 MPa respectively.
Using the yield stress calculated from tensile tests, and referring to the Eq.~\ref{eq_beam}, we calculated the maximum allowable load $F_{max}$ that can be applied to the cantilever beam before it undergoes plastic deformation. Having $F_{max}$, we then calculated the maximum allowable deflection $\delta_{max}$ and angle $\phi_{max}$  using 
\begin{align}
F_{max} &= (\sigma_{max} \times b \times t^3 ) / (6L) \label{eq_beam} \\
\delta_{max} &= (F_{max}\times x^2 \times l) / (2EI) \label{eq_beam1} \\
\phi_{max} &= (F_{max} \times x \times l) / (EI) \label{eq_beam1} .
\end{align}
The maximum allowable angle of deflection in order to remain in the linear region of the stress-strain curve is 1.96 degrees. The variation in joint angles in real laminate mechanisms are much higher than this value. Therefore, the material used as the flex layer undergoes nonlinear deformation which is subsequently considered as a source of error in the analysis of the system.

\subsection{Motion Capture}\label{sec:motion-capture}
Motion of the mechanism was recorded using two OptiTrack Prime 17W  motion tracking cameras at 360 fps rate. 
In order to sample data from mechanism oscillation, the mechanism was positioned in a known initial configuration, released, and allowed to settle into a final static configuration.
Stiffness $k$ and damping coefficients $b$ were extracted by fitting a model to the recorded data.  
We used damped natural oscillations of a simple pendulum to extract damping coefficient and stiffness at its single hinge. The design of the hinge is varied over several design parameters such as length and width of the hinge. Also cross-sectional area of the oscillating bodies were varied to study the effect of the air damping.
Based on the following model, the equation of motion of a simple pendulum made of a laminate body and a simple hinge can be written as:
\begin{align}
(I_G + m r^2)\ddot{\theta} = -k \theta - b \dot{\theta} - m g \text{sin}(\theta) \label{eq_model}
\end{align}
where $\theta$, $\dot\theta$ and $\ddot\theta$ are provided by experimental data  and $I_G$, $m$ and $r$  are extracted from models in popupCAD. A least-squares identification method was used in order to obtain the values of $k$ and $b$. 

While sampling experimental data from the single pendulum was straightforward, the data provided by the cameras for this case study demands additional data processing.  Motion-tracking cameras require three markers to be mounted to each rigid body. Thus, in order to get the information of position and orientation all the bodies, 18 markers were used to determine the 6 rigid bodies of the mechanism. Marker position and mass information was added to popupCAD as an additional 2D material layer in order to account for the added inertia.

The raw orientation data provided by the cameras were based on quaternions. In other words, data received from the cameras and Motive SDK were represented as unit quaternions for each rigid body. Since the joint axes for several of the bodies change with respect to the base frame, standard quaternion operations were used to retrieve the angle and axis of each joint over time and compared against the simulation~\cite{quaternion}.  This was done in order to produce consistent results across the simulation -- which reported state variables as the angles between adjacent bodies -- and the experiment -- which reported the quaternions of each body with respect to the base frame.

The relative quaternion between two adjacent reference frames ($q_{D_i}$) has the following relationship with quaternions of $i$th body ($q_i$) and ($i+1$)th body ($q_{i+1}$): 
\begin{align}
\quad q_{D_i} = q_{i+1} * q_i^{-1}
\end{align}
where, ($*$) is quaternion multiplication~\cite{quaternion}. 

As implemented in code, the obtained angle of rotation  using the formulation of ~\cite{quaternion} is always positive and did not change sign when the relative frame displacement went negative; instead the axis of rotation changed direction. Therefore, in order to achieve a consistent axis and angle of rotation across all time, the direction of the axis was continually monitored so that when the direction of the vector flipped, the rotation angle was negated too.


\section{RESULTS AND DISCUSSION}\label{sec:results}
\subsection{Identification of damping coefficient \& stiffness}

Damping and stiffness are due to the tension and compression of the hinge during oscillation. As a result, hinge width and length play an important role on these parameters. For damping, as well as material damping ($b_m$), drag experienced by the body increases the overall damping $b$ of the system, slowing the system and reducing oscillations faster than when in a vacuum. As air damping can be affected by the cross-sectional area of moving links, it is an important phenomenon to model.
A Fourier Series (FS) was fitted to each oscillation output and analytic derivations on the fitted FS were applied in order to obtain the angular velocity and angular acceleration of the pendulum. This method is preferred due to the amplification of noise typically found in derivatives of digital position data.
Since the test was conducted in air, drag from the moving body added to the damping of the system. In order to achieve a more precise model, both the material damping $b_m$ and air $b_a$ damping was taken in account and the overall damping of the hinges $b$ was calculated as $b = b_m + b_a$. 

\begin{figure}[tb]
\includegraphics[scale=0.5]{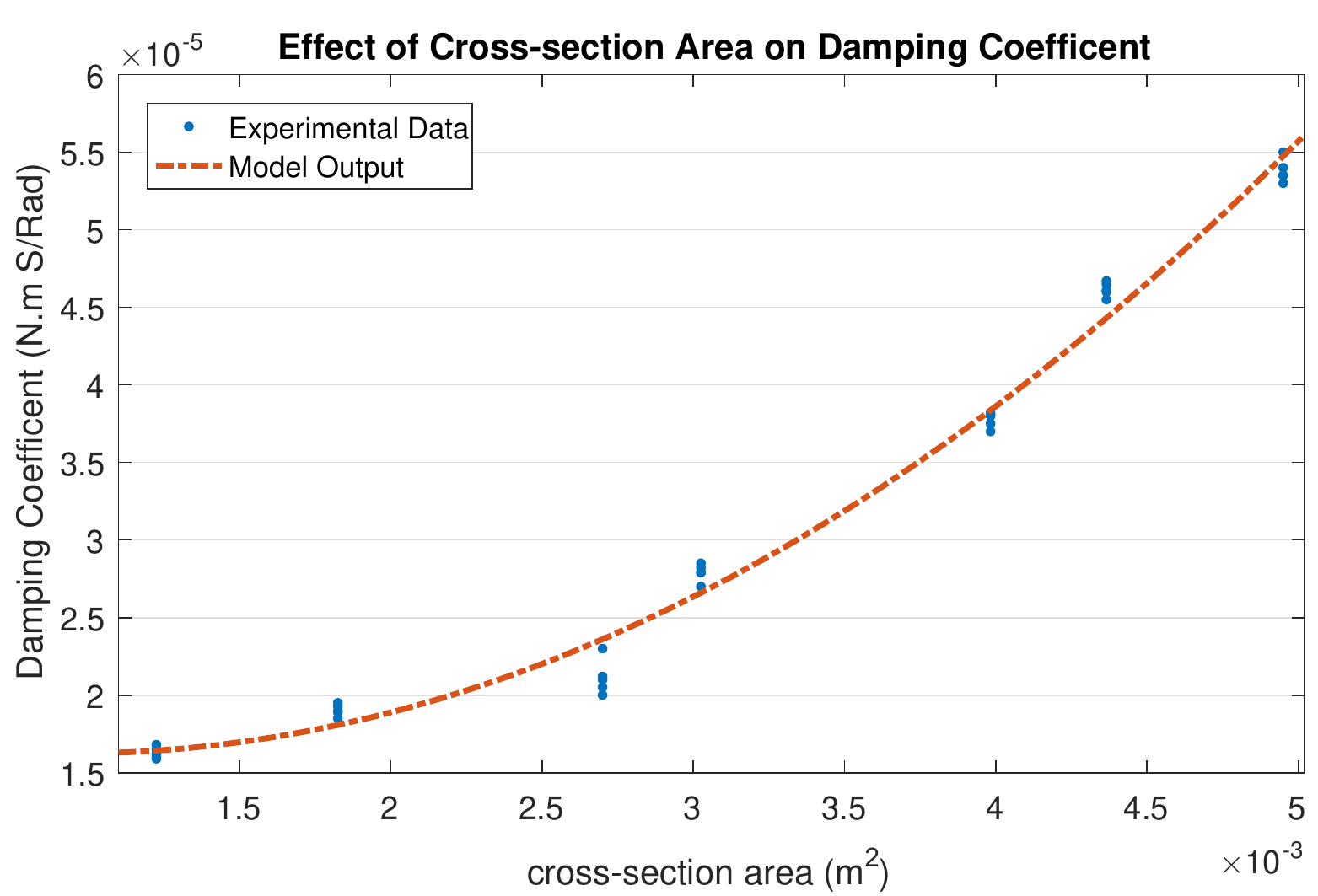}
\caption{The effect of different cross-sectional area on damping coefficient.}
\label{pic_A_On_C}
\end{figure}
The effect of air damping was studied by using seven designs where moving bodies had constant hinge designs but differing cross-sectional area. The effect of changing mass due to cross sectional area was accounted for by the $m$ fed into the simulation for each design. Figure~\ref{pic_A_On_C} depicts the change in the damping coefficient when the cross-sectional area of the moving body is changed. Based on the data, by decreasing cross-sectional area, damping coefficient also decreases. 
It should be mentioned that the authors also studied the addition of a velocity-squared term to Eq.~\ref{eq_model} in order to model damping caused by air drag. However, in the this case,the identified coefficient was negligible comparing to other coefficients. On the other hand, the coefficient of velocity term was changing as a function of cross-sectional area of the moving bodies. As a result, the velocity square term is omitted and both the air and material damping coefficients are embedded in a linear velocity coefficient. The obtained value for embedded air damping coefficient keeps the nonlinear behavior and is a second order model of cross sectional area.
\begin{align}
b = 2.34 \times a^2 - 0.0042 \times a+1.8 \times10^{-5} 
\end{align}
$1\mathrm{e}{-10}$
\subsubsection*{Effect of Hinge Width.}
In order to study the effect of hinge width on the value of  $k$ and $b$, five specimens with same body design (same cross-sectional area) were built and experimentally tested.  In order to minimize the effect of torsion across a wide range of hinge widths, smaller-width hinges were designed with gaps in the middle and constant exterior dimensions across all designs.
\begin{figure}[tb]
\includegraphics[scale=0.5]{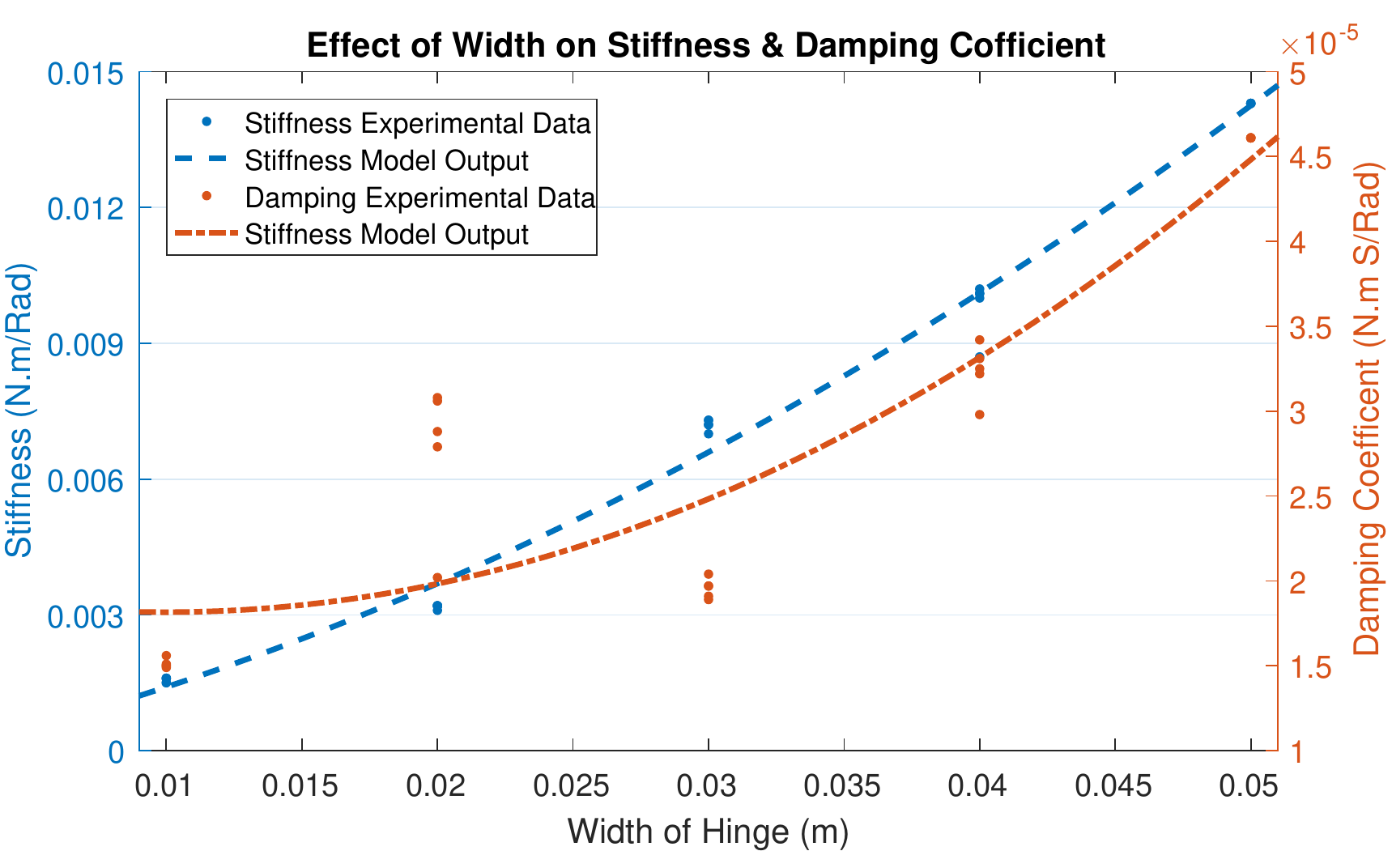}
\caption{Effect of hinge width on damping coefficient and stiffness of the hinge.}
\label{pic_NW_Effect}
\end{figure}
Figure~\ref{pic_NW_Effect} depicts the effect of the hinge width on $k$ and $b$. The second order model for damping and stiffness is given by
\begin{align}
&b = 0.0166 \times w^2 + - 3.3197 \times 10^{-4} \times w + 1.9812 \times 10^{-5}\\
&k = 3.0857 \times w^2 - 0.1361 \times w + 0.0003
\label{Eq_NW_Effect}
\end{align}

Based on the obtained results, an increase in the width of the hinge will increase the damping coefficient and stiffness of the hinge.

\subsubsection*{Effect of Hinge Length.}
\begin{figure}[tb]
\includegraphics[scale=0.5]{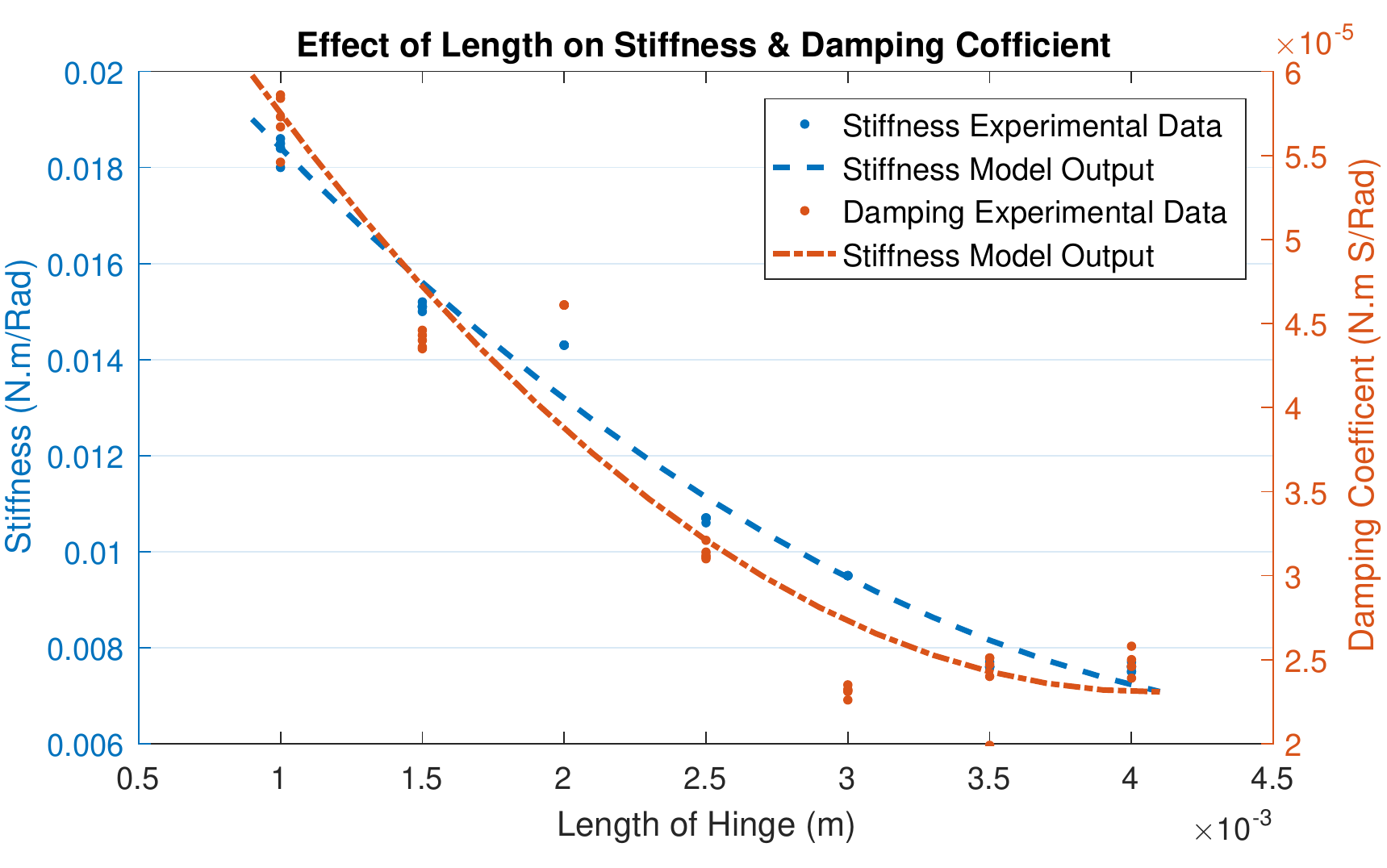}
\caption{Effect of hinge length on damping coefficient and stiffness of the hinge.}
\label{pic_NL_Effect}
\end{figure}

The effect of hinge length on $k$ and $b$ was studied across five specimens where the total hinge length was increased (Fig.~\ref{pic_NL_Effect}) which resulted in second order models given by
\begin{align}
&b = 3.6381 \times l^2 -0.0297 \times l + 8.3506 \times 10^{-5}\\
&k =  746.6667 \times l^2 -7.4590 \times l + 0.251
\label{Eq_NL_Effect}
\end{align}

The obtained results show a decrease in the value of the damping coefficient and stiffness as length increases.
\subsubsection*{Comprehensive model.}

\begin{figure}[tb]
\centering
\subfigure[Performance of identified Model for the hinge damping coefficient.]{\includegraphics[scale=0.5]{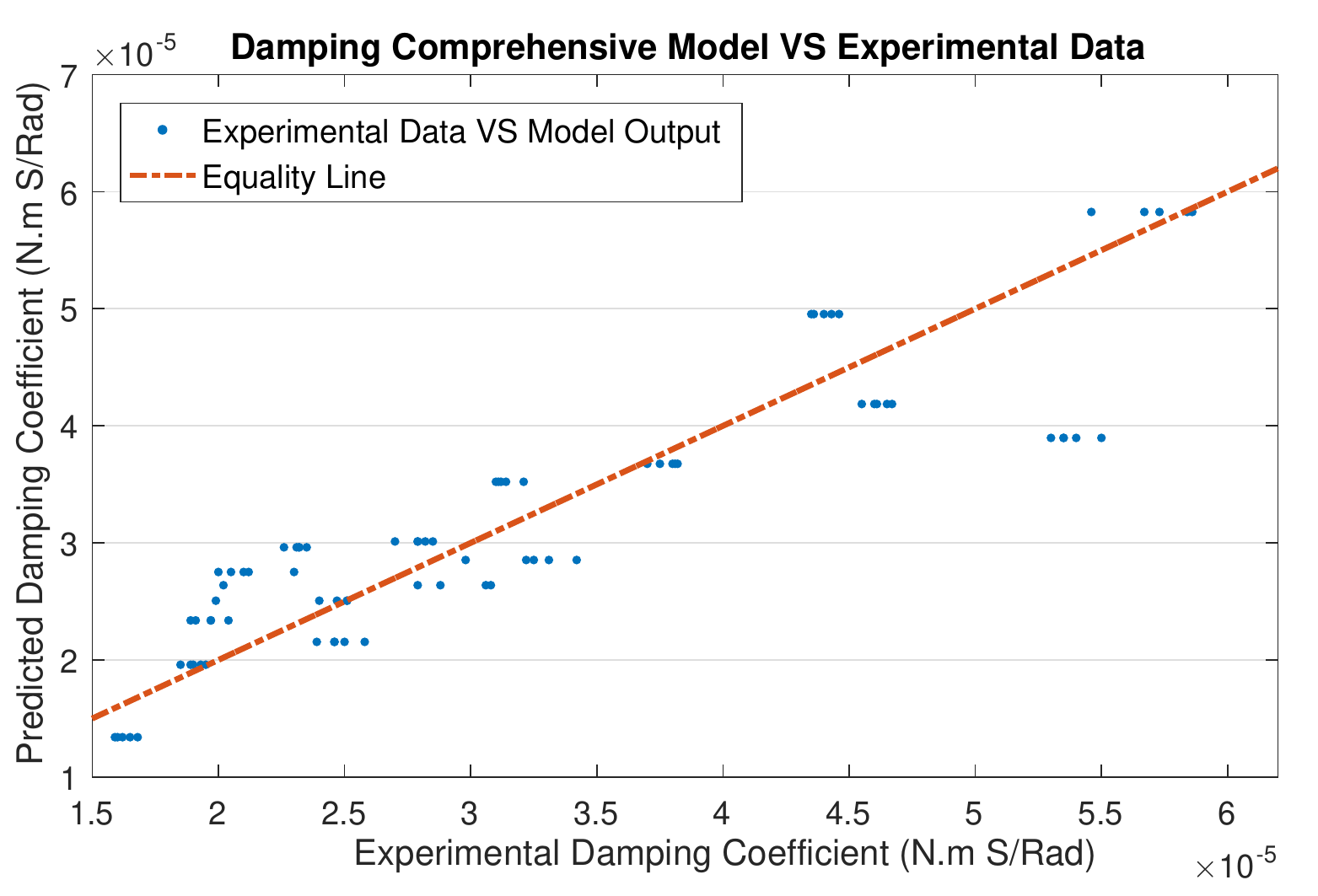}
\label{pic_All_On_C}
}
\subfigure[Performance of identified Model for the hinge stiffness.]{\includegraphics[scale=0.5]{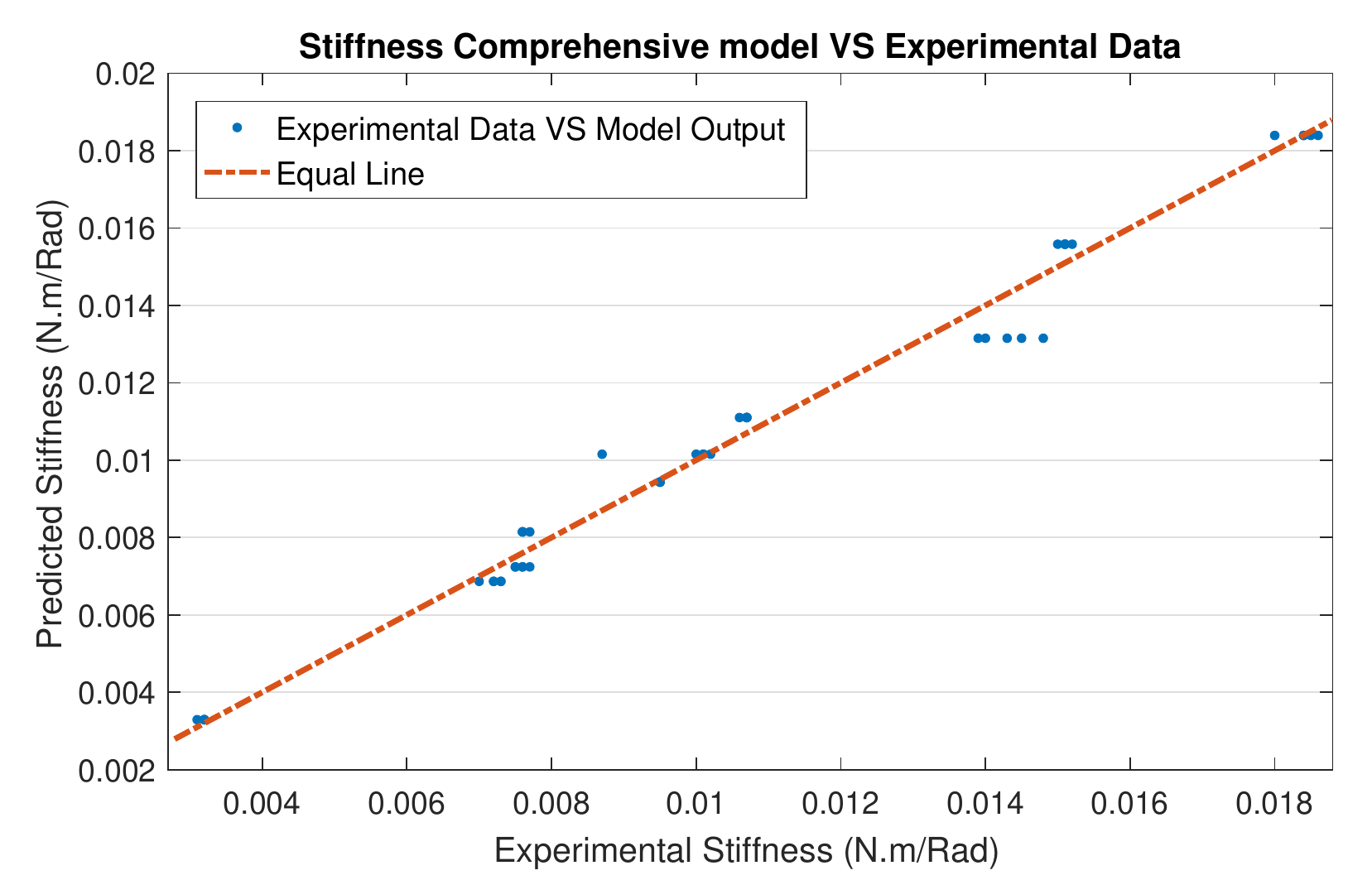}
\label{pic_All_On_K}
}
\caption{Comparison of the experimental and modeled hinge stiffness and damping coefficient}
\end{figure}

A comprehensive model was developed which takes into account all three variables ($l$,$w$, and $a$), permitting one to estimate joint properties throughout a three-dimensional design space.
Figures~\ref{pic_All_On_C} and \ref{pic_All_On_K} depict the predicted $k$ and $b$ values vs the experimental values obtained for all specimens.
\begin{align}
b = &-0.8565 \times a^2 + 0.0129 \times a + 2.0822 \times l^2 - 0.227 \nonumber \\
&  \times l + 0.0408 \times w^2 - 0.0023 \times w + 5.0855 \times 10^{-5} \label{Eq_ALL_On_C} \\
k = &  762.5397 \times l^2 - 7.5305 \times l - 1.4444 \times w^2 + \nonumber \\
& 0.4298 \times w + 0.0073 \label{Eq_ALL_On_K}
\end{align}
The obtained model for $b$ has $2.39\times10^{-6}$ ($4.01\%$) as the Mean Absolute Error (MAE), while $k$ values obtained model has $2.59\times10^{-4}$ ($1.39\%$) as MAE.  

\begin{figure*}[t]
\centering
\includegraphics[scale=0.4]{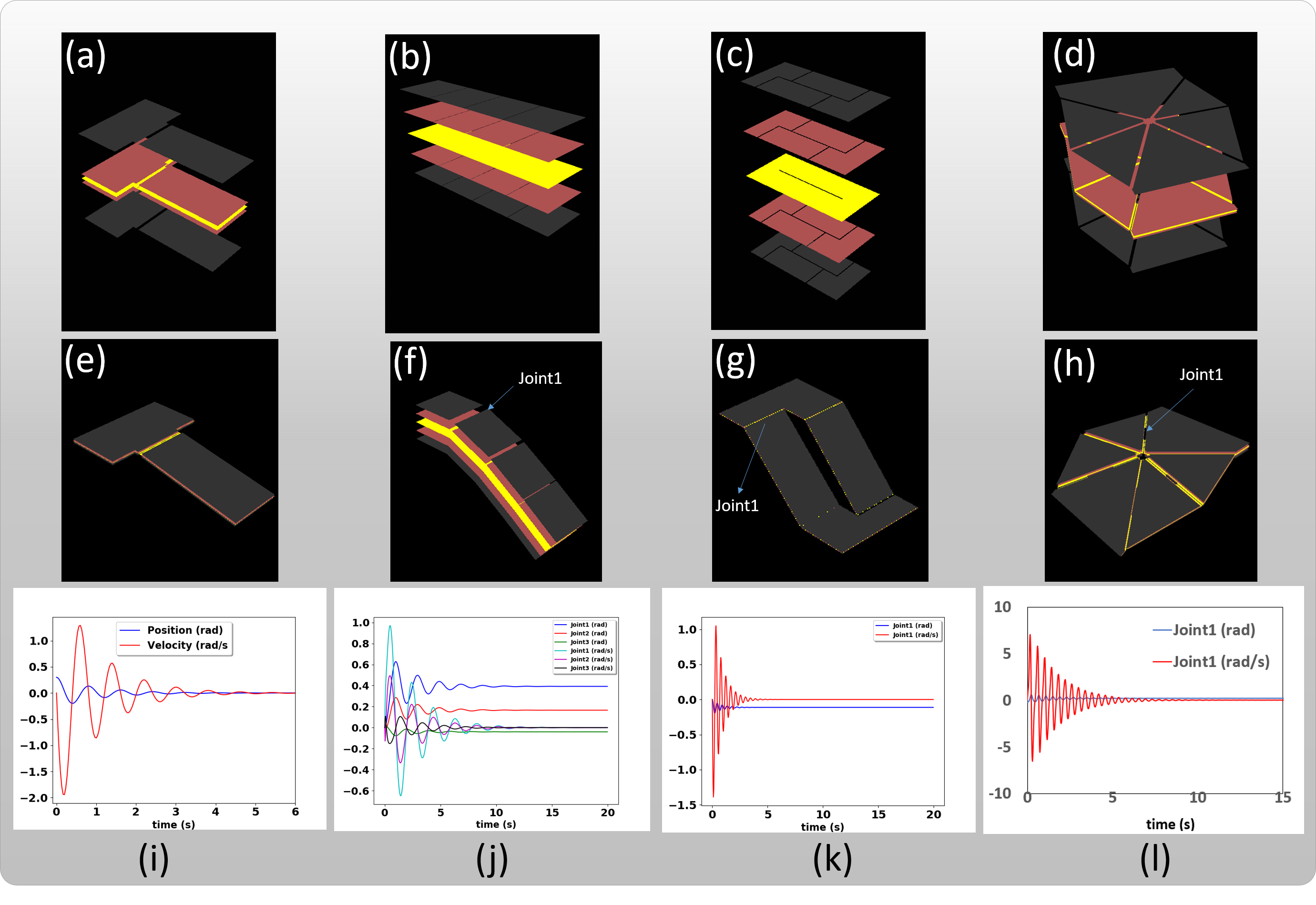}
\caption{Top row: the popupCAD design of four laminate mechanisms studied in this paper with increase in design complexity from left to right. All cases are 5 layer designs. (a) simple pendulum, (b) are a triple pendulum, (c) a 4-bar mechanism w	ith a single closed loop and (d) a 6-bar mechanism with a single closed loop, respectively. Middle row: final equilibrium position of each mechanism as predicted by Pynamics. Bottom row: the joint angles and angular velocities for some of the joints labeled in the second row as predicted by Pynamics.}
\label{pic_verification}
\end{figure*}

\subsubsection*{Model Validation.}\label{subsubsec:verification}
To test the capabilities of our integrated design and simulation environment, we have modeled a variety of laminate mechanisms in popupCAD and analyzed them using Pynamics. Figure.~\ref{pic_verification} (a)-(d)  shows the popupCAD design of four laminate mechanisms studied in this paper. All mechanisms are made of 5 layers. The complexity of the mechanisms increase from left to right. Figure.~\ref{pic_verification}(a) is the simple pendulum used for characterizing the hinges and was introduced in Section~\ref{sec:hinge-characterization}. Figure.~\ref{pic_verification}(b)-(d) are a triple pendulum, a 4-bar mechanism with a single closed loop and a 6-bar mechanism with a single closed loop, respectively. Second row of Fig.~\ref{pic_verification} shows each mechanism in its final equilibrium position as predicted by Pynamics. The joint angles and angular velocity for some of the joints labeled in the second row are plotted in the third row of  Fig.~\ref{pic_verification}(e)-(h). 
In order to evaluate how close are the positions and velocities predicted by Pynamics, to the position and velocity of the actual laminate mechanisms, we selected the most complex case in Fig.~\ref{pic_verification} as an example. 

\begin{figure}[tb]
\centering
\subfigure[Pynamics performance regarding $\theta_5$ of the 6-bar mechanism~(Fig.~\ref{pic_baumgarte}).]{
\label{pic_T6}
\includegraphics[scale=0.5]{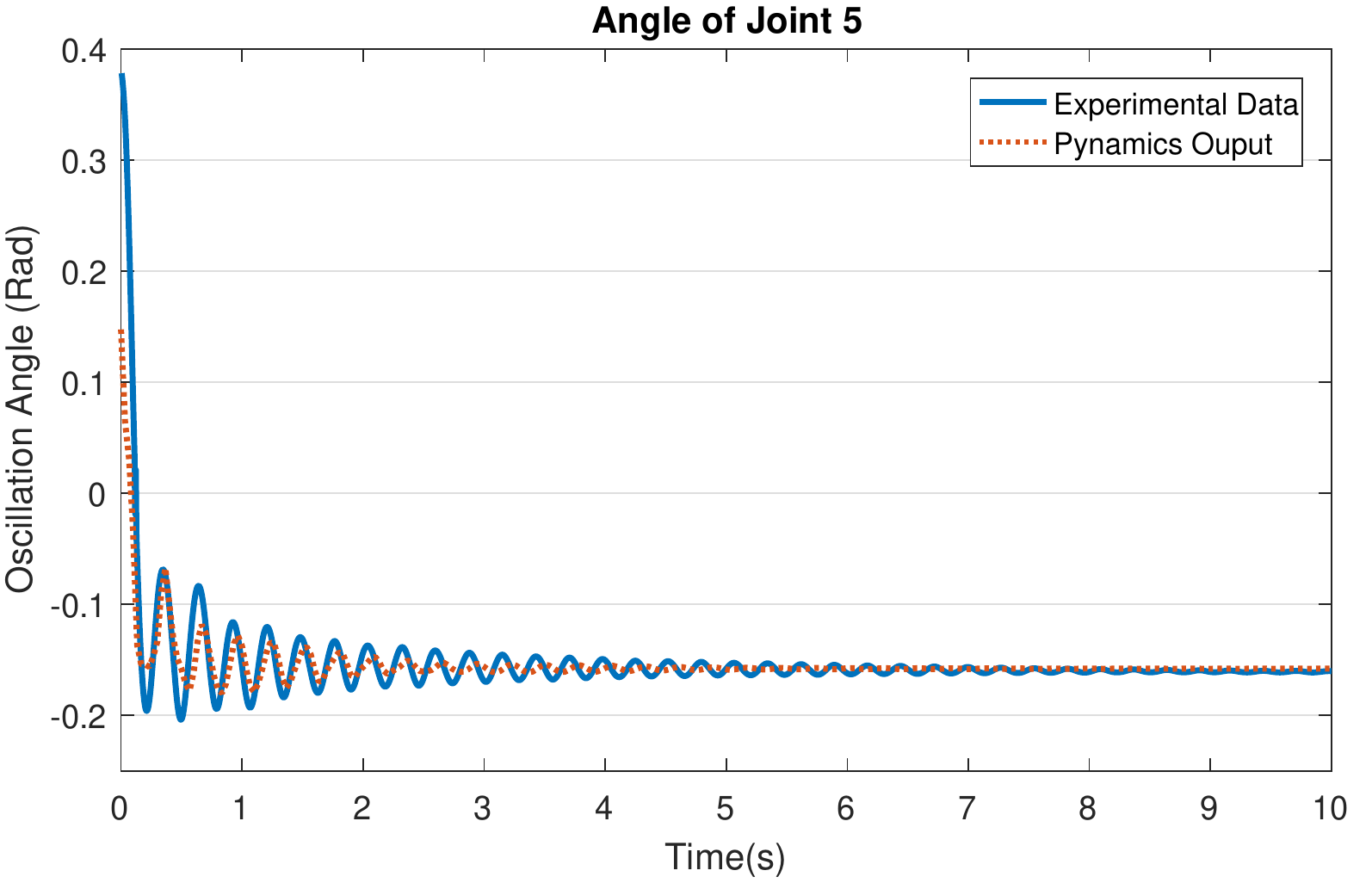}}
\subfigure[Pynamics performance regarding $\theta_6$ of the 6-bar mechanism~(Fig.~\ref{pic_baumgarte}).]{
\label{pic_T5}
\includegraphics[scale=0.5]{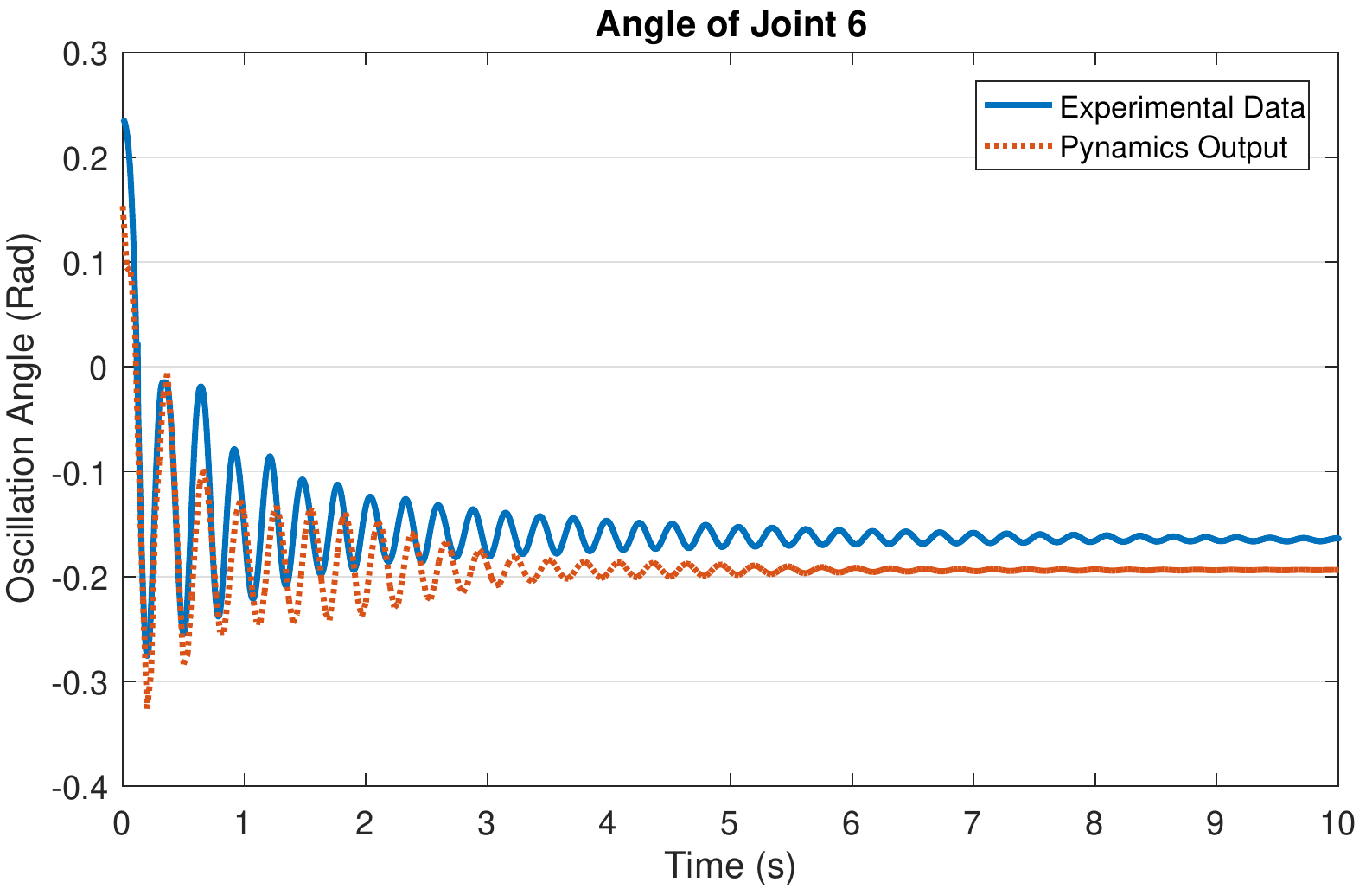}}
\caption{Pynamics output verification based on 6-bar mechanism.}
\label{pic_t5t6}
\end{figure}

Figure~\ref{pic_t5t6} illustrates the comparison between simulated and experimentally-recorded joint angles attached to body 4 of the mechanism presented in Fig .\ref{pic_baumgarte}. Despite some error, the simulation was able to properly predict the dynamic behavior of the end effector.  In particular the amplitude and wavelength of oscillations were well predicted for $\theta_5$ and $\theta_6$. Equilibrium angles obtained from simulation were also in good agreement with the experimental data as shown in Tab.~\ref{table_equil}.  Small differences between modeled and experimental equilibrium configurations can be seen in this table, as well as in the dynamic plots in Figure~\ref{pic_t5t6}.  This is partly due to our experimental method -- we used dynamic data to estimate damping and stiffness rather than steady-state force/displacement balances.  In addition, material creep, non-uniform material thickness, viscoelastic behavior, and operating in the non-linear part of the stress/strain curve are other significant reasons why this deviation may be present.
\begin{table}[tb]
\caption{The equilibrium position of Pynamics output and Experimental results in 6-bar mechanism}
\label{table_example}
\begin{center}
\begin{tabular}{c c c }
\hline
Joint & Pynamics (rad) & Experimental data (rad)\\
\hline
$\theta_1$ &  0.280 &  0.398\\
\hline
$\theta_2$ &  0.056 &  0.113\\
\hline
$\theta_3$ &  0.431 &  0.502\\
\hline
$\theta_4$ & -0.470 &  -0.737\\ 
\hline
$\theta_5$ & -0.157 &  -0.161\\
\hline
$\theta_6$ & -0.193 &  -0.166\\
\hline
\end{tabular}
\label{table_equil}
\end{center}
\end{table}
\section{CONCLUSIONS}\label{sec:conclusions}

We have developed a suite of tools which permits the dynamic simulation of laminate mechanisms by solving several challenges common across laminate devices. The resulting code integrates well with existing tools, now making it possible for a novice user to easily and quickly generate manufacturing files that are ready to be sent to machining tools with assurance that a design will work. This can help eliminate time consuming and expensive prototyping trials which are otherwise needed to validate designs without the proper analysis. While we have verified our simulations by comparing them to experimental results obtained from a simple pendulum and a spherical 6-bar mechanism, a variety of other open and closed loop designs can been simulated as seen in the accompanying video\footnote{\url{http://idealab.asu.edu/foldable_dynamics_video}}. This verifies that the automatic generation of closed loop mechanism dynamics is feasible and accurate across a wide variety of designs that are extracted from sketched mechanisms in popupCAD. We are not trying to provide a substitution to commercial software available for dynamic simulations in general, therefore, any comparison between our simulation results and those from commercial packages is not necessary.

We have characterized the hinges that are an essential part of laminate mechanisms and extracted the stiffness and damping parameters which are needed for our dynamic simulation environment. Our characterization method is simple and repeatable and therefore can be used to characterize other flexible materials used to manufacture the hinges. We have taken the effect of air damping in our model since many of the laminate mechanisms are operated in  air and this makes our simulation results closer to real working conditions of these mechanisms. The hinge parameters can be extrapolated to any new hinge design with different length, width and cross-sectional area.

Our simulation environment automates the task of generating symbolic equations of motion, integrating and solving those equations, and generating numerical and visual output. We have considered laminate mechanisms consisting of serial chains and one closed loop. Future work will include mechanisms with more than one closed loop, topological optimization of kinematics for faster simulation, considering contact between mechanisms and the ground, as well as making it possible to integrate closed-loop controllers into the simulation.

While the use of Baumgarte's method to eliminate initial-value errors was effective, it requires manual selection of $\alpha$ and $\beta$ terms.  We hope to migrate to techniques which do not require manual tuning in order to eliminate parameter selection in the future.

Material tests showed that we were using flexure material outside of its linear regime.  Further testing is required to determine if the stresses involved would lead to premature failure of these hinges.  Future mechanisms may resort to fabrics or thinner materials to minimize stresses in flexure hinges.

With those future improvements, we see this tool being used to assist novice robot designers by optimizing suggested kinematics based on stated performance goals which are then confirmed through simulation.  Ultimately, we hope this tool helps to connect design and analysis for novices to make it possible for them to design, simulate, and prototype complex robots for tasks in unstructured environments.

\bibliographystyle{asmems4}



\bibliography{asme2e}

\begin{thebibliography}{10}

\bibitem{Baisch2014}
Baisch, A.~T., Ozcan, O., Goldberg, B., Ithier, D., and Wood, R.~J., 2014.
\newblock ``{High speed locomotion for a quadrupedal microrobot}''.
\newblock {\em The International Journal of Robotics Research}, may.

\bibitem{Ma2013a}
Ma, K.~Y., Chirarattananon, P., Fuller, S.~B., and Wood, R.~J., 2013.
\newblock ``{Controlled Flight of a Biologically Inspired, Insect-Scale
  Robot}''.
\newblock {\em Science, \textbf{ 340}}(6132), may, pp.~603--607.

\bibitem{Aukes2014}
Aukes, D.~M., Goldberg, B., Cutkosky, M.~R., and Wood, R.~J., 2014.
\newblock ``{An analytic framework for developing inherently-manufacturable
  pop-up laminate devices}''.
\newblock {\em Smart Materials and Structures, \textbf{ 23}}(9), sep,
  p.~094013.

\bibitem{Aukes2014a}
Aukes, D.~M., and Wood, R.~J., 2014.
\newblock ``{Algorithms for Rapid Development of Inherently-Manufacturable
  Laminate Devices}''.
\newblock In ASME 2014 Conference on Smart Materials, Adaptive Structures and
  Intelligent Systems, ASME, p.~V001T01A005.

\bibitem{Aukes2014c}
Aukes, D.~M., Ozcan, O., and Wood, R.~J., 2014.
\newblock ``{Monolithic Design and Fabrication of a 2-DOF Bio-Inspired Leg
  Transmission}''.
\newblock In {\em Third International Conference, Living Machines 2014, Milan,
  Italy, July 30 - August 1, 2014.} Springer International Publishing, Milan,
  pp.~1--10.

\bibitem{Stellman2005}
Stellman, P., Arora, W., Takahashi, S., Demaine, E.~D., and Barbastathis, G.,
  2005.
\newblock ``{Kinematics and Dynamics of Nanostructured Origami™}''.
\newblock In Design Engineering, Parts A and B, Vol.~2005, ASME, pp.~541--548.

\bibitem{Tachi2009}
Tachi, T., 2009.
\newblock ``{Simulation of Rigid Origami}''.
\newblock In Origami4:Proceedings of 4OSME.

\bibitem{Schenk2011}
Schenk, M., and Guest, S.~D., 2011.
\newblock ``{Origami Folding : A Structural Engineering Approach}''.
\newblock In Origami 5: Fifth International Meeting of Origami Science,
  Mathematics, and Education.(5OSME), pp.~291--303.

\bibitem{Fuchi2014}
Fuchi, K., Buskohl, P.~R., Joo, J.~J., Reich, G.~W., and Vaia, R.~A., 2014.
\newblock ``{Topology Optimization for Design of Origami-Based Active
  Mechanisms}''.
\newblock In Proceedings of the ASME 2014 International Design Engineering
  Technical Conferences {\&} Computers and Information in Engineering
  Conference (IDETC/CIE 2014), pp.~DETC2014--35153.

\bibitem{Ma2013}
Ma, R.~R., Odhner, L.~U., and Dollar, A.~M., 2013.
\newblock ``{A Modular, Open-Source 3D Printed Underactuated Hand}''.
\newblock In 2013 IEEE International Conference on Robotics and
  Automation(preprint), IEEE.

\bibitem{Doshi2015c}
Doshi, N., Goldberg, B., Sahai, R., Jafferis, N., Aukes, D., Wood, R.~J., and
  Paulson, J.~A., 2015.
\newblock ``{Model driven design for flexure-based Microrobots}''.
\newblock In 2015 IEEE/RSJ International Conference on Intelligent Robots and
  Systems (IROS), IEEE, pp.~4119--4126.

\bibitem{Hanna2014}
Hanna, B.~H., Lund, J.~M., Lang, R.~J., Magleby, S.~P., and Howell, L.~L.,
  2014.
\newblock ``{Waterbomb base: a symmetric single-vertex bistable origami
  mechanism}''.
\newblock {\em Smart Materials and Structures, \textbf{ 23}}(9), sep,
  p.~094009.

\bibitem{Aukes2015}
Aukes, D.~M., and Wood, R.~J., 2015.
\newblock ``{PopupCAD: a tool for automated design, fabrication, and analysis
  of laminate devices}''.
\newblock In SPIE.DSS, T.~George, A.~K. Dutta, and M.~S. Islam, eds.,
  p.~94671B.

\bibitem{Baumgarte1972}
Baumgarte, J., 1972.
\newblock {Stabilization of constraints and integrals of motion in dynamical
  systems}.

\bibitem{masarati2011adding}
Masarati, P., 2011.
\newblock ``Adding kinematic constraints to purely differential dynamics''.
\newblock {\em Computational Mechanics, \textbf{ 47}}(2), pp.~187--203.

\bibitem{ASTMInternational2012}
{ASTM International}, 2012.
\newblock ``{ASTM D882: Standard Test Method for Tensile Properties of Thin
  Plastic Sheeting}''.
\newblock {\em ASTM Standards}, p.~12.

\bibitem{quaternion}
Aydin, Y., and Kucuk, S., 2006.
\newblock ``Quaternion based inverse kinematics for industrial robot
  manipulators with euler wrist''.
\newblock In Mechatronics, 2006 IEEE International Conference on, IEEE,
  pp.~581--586.

\end{thebibliography}



 \end{document}